# ARASPIDER: DEMOCRATIZING ARABIC-TO-SQL

Ahmed Heakl[1], Youssef Mohamed[2*], and Ahmed B. Zaky[3]

[1]Department of Computer Science, Egypt-Japan University of Science and Technology
ahmed.heakl@ejust.edu.eg
[2]Department of Computer Science, Egypt-Japan University of Science and Technology
youssef.khalil@ejust.edu.eg
[3]Department of Computer Science, Egypt-Japan University of Science and Technology
ahmed.zaky@ejust.edu.eg

## ABSTRACT

This study presents AraSpider, the first Arabic version of the Spider dataset, aimed at improving natural language processing (NLP) in the Arabic-speaking community. Four multilingual translation models were tested for their effectiveness in translating English to Arabic. Additionally, two models were assessed for their ability to generate SQL queries from Arabic text. The results showed that using back translation significantly improved the performance of both ChatGPT 3.5 and SQLCoder models, which are considered top performers on the Spider dataset. Notably, ChatGPT 3.5 demonstrated high-quality translation, while SQLCoder excelled in text-to-SQL tasks. The study underscores the importance of incorporating contextual schema and employing back translation strategies to enhance model performance in Arabic NLP tasks. Moreover, the provision of detailed methodologies for reproducibility and translation of the dataset into other languages highlights the research's commitment to promoting transparency and collaborative knowledge sharing in the field. Overall, these contributions advance NLP research, empower Arabic-speaking researchers, and enrich the global discourse on language comprehension and database interrogation.

## 1. INTRODUCTION

The limitations of existing semantic parsing datasets, characterized by simplistic SQL queries and homogeneous database schemas, underscore the pressing need for more diverse and complex datasets to advance the field. The challenge of generalizing semantic understanding to unseen SQL queries and database structures remains a critical barrier to achieving robust performance in text-to-SQL conversion tasks. The introduction of the Spider dataset [1] addresses these challenges, providing a large-scale, diverse corpus that presents a formidable testbed for evaluating and improving semantic parsing and text-to-SQL conversion models. The Spider dataset stands out as a groundbreaking resource due to its expansive collection of 200 databases spanning 138 domains, each containing multiple tables with complex relationships, enabling researchers to evaluate model performance across diverse scenarios. Its innovative train-test split, wherein different databases and SQL queries are employed in each set, ensures rigorous evaluation of models' generalization capabilities to unseen data, a crucial aspect often overlooked in previous datasets. Researchers can leverage the intricacies of Spider to not only benchmark their semantic parsing and text-to-SQL models but also to push the boundaries of natural language understanding in complex database querying tasks, ultimately advancing the state-of-the-art in NLP research.

Motivated by the considerable presence of Arabic speakers worldwide, with 109.66 million individuals as native speakers and a broader audience of 372.7 million, this research aims to bridge a critical gap in the natural language processing (NLP) field by providing a comprehensive translation of the Spider dataset into the Arabic language. Building upon recent advancements in multilingual text-to-SQL translation, such as [2, 3], which have extended the accessibility of database querying tasks to various linguistic communities, this paper seeks to enhance the inclusivity and applicability of semantic parsing methodologies within the Arabic-speaking community. By extending the accessibility of the Spider dataset to Arabic speakers, this research endeavors to foster a more robust and diverse landscape for NLP research, empowering Arabic-speaking researchers and practitioners to contribute meaningfully to the advancement of semantic parsing and text-to-SQL conversion technologies. Through this endeavor, we aspire to not only democratize access to cutting-edge NLP resources but also to catalyze innovation and collaboration within the Arabic NLP community, ultimately enriching the global discourse on natural language understanding and database querying.

In addition to the aforementioned endeavors, this research makes four significant contributions to the field of natural language processing and semantic parsing. Our contributions are as follows:

- To the best of our knowledge, AraSpider is the first Arabic version of the Spider dataset.
- We proposed an extensive evaluation of 4 multilingual translation models to identify the best model for translating NL questions.
- We evaluated Text-to-SQL across top-performing language models on the Spider dataset.
- We provided a pipeline for non-English-to-SQL without finetune on non-English questions.
- We provided written scripts for reproducing results and translating the dataset for other languages to assist future research.[1]

## 2. RELATED WORKS

The field of Text-to-SQL has been actively researched in recent years, as evidenced by various datasets and benchmarks [7, 8, 15, 16, 17, 6, 19]. The Spider benchmark is the most famous and widely used. It is specifically designed to create natural language interfaces for databases spanning different domains. This benchmark comprises 10,181 questions and 5,693 unique complex SQL queries distributed across 200 databases with multiple tables, encompassing 138 diverse domains. In Spider 1.0, both the training and test sets feature a variety of complex SQL queries and databases. Achieving high performance on this benchmark requires systems to exhibit robust generalization capabilities, handling not only new SQL queries but also adapting to novel database schemas.

Several efforts have been made to extend the Spider dataset to other languages, such as the creation of Chinese Spider (CSpider) [20]. CSpider involves translating original questions from the Spider dataset from English to Chinese, incorporating a specific cultural localization of certain values. It is important to note that this translation is not a literal one, and while the principles of localization were applied, they were not explicitly detailed. The authors conducted a comparison between models trained on machine-translated questions and those trained on NL queries translated manually with subsequent localization. The findings indicate that models trained on human translation exhibit significantly better performance than those relying on machine translation. This underscores the importance of professional translation and localization when adapting English datasets to other languages. However, it is noteworthy that CSpider did not introduce new localized values into the database content, limiting a comprehensive evaluation of the results obtained.

---
[1] Codebase at github.com/ahmedheakl/AraSpider

Three other specific variations of the Spider dataset include the Vietnamese version [21], the Portuguese version [2], and the MultiSpider dataset [14]. In the case of the Vietnamese version, both NL questions and the database schema, encompassing table and column names, as well as values in SQL queries, were translated. Unlike the Chinese Spider, the translated values in the Vietnamese version were not localized and were not integrated into the database content. Following a similar experimental approach as CSpider, the study compared manually-translated and machine-translated versions of the Vietnamese dataset. In the case of the Portuguese Spider, only NL questions were translated, and no alterations affecting the database schema or content were introduced. It is evident that across all existing versions of the Spider dataset, despite variations in the principles governing language adaptation, not all components of the original dataset undergo modifications. Specifically, the database content remains unaltered and does not align with the translated questions. Consequently, the evaluation of model performance relies solely on exact matching since execution accuracy cannot be utilized due to the unchanged database content. Lastly, the work by [14] introduces significant advancements in the Text-to-SQL semantic parsing task. Focused on enhancing user-database interactions, their work presents the MultiSpider dataset, a notable contribution as the largest multilingual text-to-SQL dataset encompassing seven languages—English, German, French, Spanish, Japanese, Chinese, and Vietnamese. The authors not only addressed the scarcity of high-quality multilingual datasets but also identified and tackled lexical and structural challenges arising from specific language properties and dialect nuances. Notably, the authors introduce the Schema-Augmentation-with-Verification (SAVe) framework, a simple yet effective data augmentation method from a schema perspective. Through extensive experiments, they demonstrate the challenges posed by MultiSpider and showcase how SAVe significantly enhances overall performance, thus closing the performance gap across languages by 29.5%. The paper's contributions lie in providing a comprehensive multilingual dataset, identifying linguistic challenges, and proposing a novel schema-centric augmentation approach, shedding light on the intricacies of multilingual text-to-SQL semantic parsing.

Current methods in Natural Language to SQL (NL2SQL) can be categorized into two main types: entity-based and machine-learning approaches, with the latter being predominantly driven by deep learning techniques [1]. Entity-based methodologies concentrate on interpreting input text using predefined rules to transform it into an SQL query. Typically, this translation occurs in two stages: first to an intermediary state and then to the final SQL query. Notable systems employing entity-based approaches include Bela [4], SODA [5], NaLIR [6], TR Discover [9], Athena [10], Athena++ [12], and Duoquest [13]. Machine learning approaches are based on supervised learning, in which training data contains natural language questions and paired SQL queries [3]. Several architectures can be trained or fine-tuned to run the translation. Relevant systems are Seq2SQL [19], SQLNet [22], RAT-SQL [23], HydraNet [24], DIN-SQL [25], DAIL-SQL [26], and SQLCoder by defog.

The previous extensive reference lists provide ample evidence that the translation from natural language to SQL has been extensively researched, signifying a mature field with the widespread use of benchmarks encompassing training and testing data, as well as evaluation methods for novel proposals in NL2SQL. However, it is less adapted to non-English languages, particularly the Arabic language.

## 3. METHODOLOGY

To provide efficient and reliable Arabic translation of the spider dataset, we opt to make sure our translation is correct and evaluate multiple models on language translation and the text-to-SQL task. This section covers the details of both steps.

## 3.1. Translation Pipeline

We present the translation pipeline we used to generate the AraSpider dataset.

### 3.1.1. Translation and Validation

To be effective, we first used ChatGPT 3.5 to translate the Spider dataset questions from English to Arabic, and then let each translator post-edit the translation individually. We used ChatGPT specifically as we are targeting the problem of the questions being only translated rather than the questions, the schema, and the database. In other words, ChatGPT can benefit from the contextual information in the schema, giving more accurate translations. We achieved so using prompt engineering, Figure 1.

> Imagine you are a translator from English to Arabic for some questions. These questions are accompanied by a database schema, please take it into context while translating. Give me the translation and the translation only of the following questions. Here are some instructions to stick to while translating:
>     - Do not translate words or phrases in quotations.
>     - Try to generate translations in question format.
> Just give me the Arabic translation of the questions, no need to translate the schema.
> > Questions and schemas:
> [Question] {question}
> [Schema] {schema}

Figure 1. Prompt for translating the Spider dataset questions from English to Arabic using ChatGPT 3.5

### 3.1.2. Qualified Translators

Inspired by the work of [14] to hire qualified translators, namely 2 of the authors of this paper have an IELTS score of >= 7.0 (C1 proficiency). Also, both are native speakers of the Arabic language, meaning they can efficiently evaluate and edit other translations. An essential component of post-edit review involves incorporating Arabic cultural nuances into the translation process. It is crucial that translations from English to Arabic transcend literal interpretations, and the involvement of native translators becomes paramount in preserving the authentic Arabic identity. This ensures the replication of real-world settings in which Arabic-to-SQL systems would be deployed.

## 3.2. Translation Evaluation

As the Spider dataset is not the only publicly available dataset, we opt to look for the best translation pipeline for future research. To achieve that endeavor, we evaluated the translation of 200 samples from the Spider dataset carefully chosen to strengthen the issues generated by most translation models. The criteria for picking these 200 random samples were as follows (1) a subset should contain sentences with words in quotations, (2) the set should contain both short and long sentences (5 ~ 20 words per question), (3) and some of the sentences should be in order format and others in question format. The reason for picking each criterion is (1) typically when words are in quotations, this means that the word/phrase between quotations should be used literally in the SQL query. Hence, with this criterion, we test the model's ability to preserve such words/phrases. Regarding (2), the translation ability of the utilized model should be diverse, i.e. it should handle both long and short translations accurately. Lastly, (3) as users can either ask the model to do something (order format) or ask the model a question (question format), the translation model should translate each case appropriately while maintaining the question format in the translation. Some samples are shown in Table 1.

Table 1. Sample questions and translations from the Spider dataset.

| Question | Translation |
| --- | --- |
| List the creation year, name and budget of each department. | اعرض قائمة بسنوات الإنشاء، وأسماء وميزانيات كل قسم. |
| What is the average number of employees of the departments whose rank is between 10 and 15? | ما هو المتوسط لعدد الموظفين في الأقسام الذين يحتلون المرتبة بين 10 و 15؟ |
| What are the distinct creation years of the departments managed by a secretary born in state 'Alabama'? | ما هي سنوات الإنشاء المميزة للأقسام التي يديرها أمين وُلد في ولاية 'Alabama'؟ |
| What are the themes of farm competitions sorted by year in ascending order? | ما هي المواضيع التي تم تنظيمها في مسابقات المزارع مرتبة حسب السنة بترتيب تصاعدي؟ |

We cherry-picked 4 models for evaluation, namely ChatGPT, Google Translate API, Opus MT Big, and Marafa MT. Noticeably, the former 2 of these models are closed source, and the latter 2 are open source. To evaluate the translations for these models, we used ChatGPT as an evaluator giving a score out of ten for each generated translation from each model. As it be noticed in Figure 2, we used few-shot prompting to instruct the model of how the input English and Arabic translation might look.

> I am providing the next sentences translated from English to Arabic. I want to evaluate the translation of each sentence for 0 to 10 where 0 is poor and 10 is perfect. Do not provide a reason or an explanation, just provide the rate as a number.
> - Sentence: For each start station id, what is its name, longitude and average duration of trips started there? : [Translation] لكل معرف محطة البداية، ما هو اسمها وخط الطول والمدة المتوسطة للرحلات التي بدأت منها؟
>
> - Sentence: For each station, find its latitude and the minimum duration of trips that ended at the station. : Translation: [Translation] لكل محطة، اعرض خط العرض الخاص بها والمدة الدنيا للرحلات التي انتهت في المحطة.
>
> Give me the rate for the following translation: {translation}
> > Rate:

Figure 2. Prompt for rating translations out of 10 using ChatGPT 3.5

### 3.3. Text-to-SQL Evaluation

To evaluate the effectiveness of our translation and explore the potential drawbacks of the task Arabic to SQL, we utilized the highest achieving models on the Spider dataset, i.e. ChatGPT and SQLCoder 7b-2. Particularly, ChatGPT is used amongst the common industrial business intelligence automation solutions such as LiDA and Vanna.ai, which are commonly used across multiple business domains. Moreover, the newest version of SQLCoder, in addition to being open source, promises higher execution accuracy than ChatGPT on multiple benchmarks. Specifically, SQLCoder 7b-2 is knowledge distilled from the SQLCoder 70b version trained based on CodeLLaMa. The evaluation of these state-of-the-art models gives the research some insights into the open-ended challenges related to Arabic-to-SQL.

We consider two forms of evaluation. Firstly, direct Arabic-to-SQL, where the Arabic-translated question from the Spider dataset is provided and the model has to generate the right SQL query, Figure 3.

> ### Examples
> Some example questions and corresponding SQL queries are provided based on similar problems:
> Answer the following: ماهى العناوين الالكترونية للمستخدمين؟
> SELECT email FROM users;
>
> ### Instructions:
> Your task is to convert a question into a SQL query, given an SQLite database schema. Adhere to these rules:
>   - **Deliberately go through the question and database schema word by word** to appropriately answer the question
>   - **Use Table Aliases** to prevent ambiguity. For example, `SELECT table1.col1, table2.col1 FROM table1 JOIN table2 ON table1.id = table2.id`.
>   - When creating a ratio, always cast the numerator as a float
>   - **Use only one column per SQL query**
>
> ### Input:
> Generate a SQL query that answers the question `{question}`.
> This query will run on a database whose schema is represented in this string:{schema}
> ### Response:
> Based on your instructions, here is the SQL query I have generated to answer the question `{question}`:
> ```sql

Figure 3. Prompt for generating SQL queries from Arabic question. The prompt is used as an input for both ChatGPT 3.5 and SQLCoder models.

The second form of evaluation is first translating the Arabic question into English using ChatGPT and then prompt the models again - we call this process back-translation-to-SQL, Figure 4.

## 4. RESULTS

> Imagine you are a translator from Arabic to English in the following text. Give me the translation and the translation only.
> > Question: {text}
> > English translation:

Figure 4. Prompt for back translation using ChatGPT 3.5.

In this section, the results of the evaluation are presented, which consist of two main assessments: translation evaluation and text-to-SQL evaluation. The translation evaluation assesses the quality of English-to-Arabic translations produced by three different large language models (LLMs), with a primary reliance on ChatGPT-3.5 as the benchmark. Subsequently, the text-to-SQL evaluation focuses on appraising the accuracy and effectiveness of the translated Arabic text in generating SQL queries, employing two LLMs for evaluation purposes.

4.1. **Translation Evaluation**

In this evaluation test, translations of English questions to Arabic questions were conducted using three LLMs, which were subsequently evaluated by ChatGPT 3.5. Among these models, two are free and open-source: Arabic Opus MT and Arabic Merafa MT, while the third is ChatGPT-3.5 itself. The evaluation criterion entails calculating the sum of scores for all sentences, dividing by the total number of sentences, and then multiplying by 100, as depicted in Eq. (1). The results for all models are presented in Table 2.

$$score = \frac{Sum\ of\ the\ scores\ of\ all\ sentences}{total\ no.of\ sentences} \times 100 \quad\quad \text{Eq. (1)}$$

Table 2. Models' average evaluation score

| Model Name | Average Score |
|---|---|
| ChatGPT 3.5 | **8.27** |
| ChatGPT 3.5 (no context) | 7.7 |
| Arabic Opus MT | 7.8 |
| Arabic Marefa MT | 6.4 |
| Google Translate | 7.7 |

As depicted in Table 2, the ChatGPT 3.5 translation exhibits the highest total score, surpassing Arabic Opus MT by 0.47 points, Arabic Marefa MT by 1.87 points and Google Translate by 0.57 points. Notably, the disparity between ChatGPT-3.5 and Arabic Opus is relatively small, which highlights an advantage for Arabic Opus MT due to its status as a free and open-source model.

4.2. **Text-To-SQL evaluation**

In this evaluation test, two LLMs were employed to convert natural language inputs into SQL queries and calculate execution accuracy. Execution accuracy is determined by comparing the predicted query's output with the gold standard query, with consideration given to multiple correct queries as advantageous. Two types of inputs were provided to the LLMs: Arabic questions and English questions, with the latter obtained through back translation of the Arabic questions. Back translation involves converting the Arabic text to English via ChatGPT-3.5. This technique is used because the LLMs used in the evaluation are not fine-tuned in Arabic text.

As demonstrated in Table 3, back translation results in a 24.3% improvement in ChatGPT-3.5 outcomes and a 23.4% enhancement in SQLCoder's performance. Moreover, the SQLCoder model exhibits a 19.5% higher accuracy than ChatGPT 3.5 in back translation scenarios and a 20.4% higher accuracy than ChatGPT 3.5 when processing Arabic text directly.

The results of the evaluation highlight the performance of different LLMs in both translation and text-to-SQL tasks. While ChatGPT 3.5 demonstrates superior translation quality compared to the other LLMs evaluated, the advantages of free and open-source models such as Arabic Opus MT are underscored by their competitive scores. In the text-to-SQL evaluation, the use of back translation significantly improves the performance of ChatGPT 3.5, showcasing its adaptability to handle Arabic text through intermediary English conversion. Additionally, the SQLCoder emerges as a robust model, outperforming ChatGPT-3.5 in both Arabic and English inputs, emphasizing its efficacy in generating SQL queries accurately.

Table 3. Execution accuracy scores of different LLMs for the text-to-SQL task were evaluated on both direct Arabic-to-SQL and back-translation-to-SQL.

| Model Name | Average Score |
| --- | --- |
| ChatGPT-3.5-Arabic | 28.5% |
| ChatGPT-3.5-Back-Translation | 52.8% |
| ChatGPT-3.5-English | 82.0% |
| SQLCoder-Arabic | **48.9%** |
| SQLCoder-Back-Translation | **72.3%** |
| SQLCoder-English | **85.4%** |

## 5. DISCUSSION

In this section, key findings and conclusions from our research and experiments are highlighted. These insights are intended to serve as valuable guidance for future studies, aiding researchers in avoiding the reinvention of the wheel and benefiting from our experimental outcomes. The experiments indicate that existing models along with back translation perform strongly, and hence omit the need to finetune such models on Arabic datasets for certain applications.

Overall, questions in English have a closer similarity with SQL queries, and hence simplify the SQL generation process. However, questions in Arabic require more understanding, which could explain why back translation boosts that execution accuracy. Additionally, our translation was accompanied by Arabic culture to mimic the real production in which Arabic-to-SQL system are used. For example, for the question "List the most popular items among young adults," has a literal translation of "اعرض قائمة بالعناصر الأكثر شعبية بين الشباب البالغين". Yet, understanding what constitutes "young adults" and what qualifies as "popular" requires cultural and contextual knowledge to generate an appropriate SQL query, which was included in our translation.

### 5.1. Common Mistakes in Translation

During the translation process, some common mistakes amongst translation were repeated. We present these mistakes below:

- Arabic numerals are often overlooked for translation, necessitating their conversion to Arabic script.
- Instances exist where text enclosed within quotation marks is inaccurately translated into Arabic.
  For example: <What are the hosts of competitions whose theme is not "Aliens"?> will be translated to <ما هي المضيفات للمسابقات التي ليست موضوعها "الكائنات الفضائية"؟>
- Translation can change the order of the sentence.
- Instances occur where translated words are inaccurately rendered as nouns rather than verbs.
  For example: <List the creation year, name and budget of each department.> will be translated to <قائمة بسنوات الإنشاء، وأسماء وميزانيات كل قسم>.

### 5.2. Using the Schema for Translation

Utilizing contextual schema enhances translation quality and aids in generating SQL queries for language models, as demonstrated in Table 2. The incorporation of context into ChatGPT-3.5 resulted in an average translation score increase of 0.57 compared to translations conducted without schema. This shows the power of large language models as compared prompts, as compared to Google Translate where only the question is provided. This is obvious from the fact that ChatGPT translation without context achieved the exact score as Google translate, which strengthens our claims.

### 5.3. Using Back Translation for Text-to-SQL Generation

Using back translation involves converting Arabic text to English before inputting the question into ChatGPT-3.5 or SQLcoder models. This approach is adopted due to the models' superior proficiency in handling English text compared to Arabic. By employing back translation without fine-tuning on Arabic text, a fair evaluation for the task is facilitated, ensuring equitable assessment of model performance. This could be explained due to the fact that databases in the Arab world are still implemented in English, i.e. the schema is in English. Hence, providing text-to-SQL models with English questions makes more sense.

### 5.4. Recommended Text-to-SQL Pipeline with No Finetuning

Given previous evaluation, we suggest that to employ text-to-SQL for other languages, the following pipeline could be used to avoid finetuning. As stated before, back translation boosts both ChatGPT and SQLCoder accuracy, and hence we suggest to first translate the original language to English. We suggest using ChatGPT for such translation given its ability to take contextual information into account. Lastly, for text-to-SQL generation, we suggest using SQLCoder model because of its strong performance and its high accuracy surpassing ChatGPT4 and ChatGPT4-turbo.

## CONCLUSION AND FUTURE WORK

In conclusion, this research introduces AraSpider, the first Arabic rendition of the Spider dataset, a pivotal step in advancing natural language processing within the Arabic-speaking community. Through a rigorous evaluation of multilingual translation models, our findings highlight the efficacy of back translation in enhancing the performance of models like ChatGPT 3.5 and SQLCoder, particularly on the challenging Spider dataset. This work not only addresses the limitations of existing semantic parsing datasets but also aims to foster inclusivity, innovation, and collaboration within the Arabic NLP community, contributing to the global discourse on language comprehension and database interrogation.

Our four key contributions, including the creation of AraSpider, an evaluation of translation models, an assessment of text-to-SQL tasks, and the provision of reproducible methodologies, collectively advance NLP research, empower Arabic-speaking researchers, and set the stage for future endeavors in semantic parsing and text-to-SQL conversion. By emphasizing the importance of contextual schema and back translation strategies, this research provides valuable insights for enhancing model performance in Arabic NLP tasks, ultimately contributing to the ongoing evolution of natural language understanding and database querying technologies.

Future work should evaluate on larger models such ChatGPT4, ChatGPT4-turbo, and SQLCoder 70b. Emergent abilities might be more obvious in these models due to their huge parameter space, allowing them to understand Arabic questions and generate SQL queries easier. Also, translation of other datasets, such as WikiSQL, could be employed using our translation pipeline. Lastly, finetuning open-source models such as SQLCoder for Arabic text-to-SQL can be implemented.

## REFERENCES

[1] Yu, T., Zhang, R., Yang, K., Yasunaga, M., Wang, D., Li, Z., ... & Radev, D. (2018). *Spider: A large-scale human-labeled dataset for complex and cross-domain semantic parsing and text-to-SQL task.* arXiv preprint arXiv:1809.08887.

[2] José, M. A., & Cozman, F. G. (2021). mRAT-SQL+ GAP: a Portuguese text-to-SQL transformer. In Intelligent Systems: 10th Brazilian Conference, BRACIS 2021, Virtual Event, November 29–December 3, 2021, Proceedings, Part II 10 (pp. 511-525). Springer International Publishing.

[3] Jose, M. A., & Cozman, F. G. (2023). *A multilingual translator to SQL with database schema pruning to improve self-attention.* International Journal of Information Technology, 1-9.

[4] Walter, S., Unger, C., Cimiano, P., Bär, D.: Evaluation of a layered approach to question answering over linked data. Lect. Notes Comput. Sci. (including Subser. Lect. Notes Artif. Intell. Lect. Notes Bioinformatics). 7650 LNCS, 362–374 (2012). https://doi.org/10.1007/978-3-642-35173-0-25

[5] Blunschi, L., Jossen, C., Kossmann, D., Mori, M., Stockinger, K.: SODA: Generating SQL for business users. Proc. VLDB Endow. 5, 932–943 (2012). https://doi.org/10.14778/2336664.2336667

[6] Li, F., Jagadish, H. V: Constructing an interactive natural language interface for relational databases. Proc. VLDB Endow. 8, 73–84 (2014). https://doi.org/10.14778/2735461.2735468

[7] Yujian Gan, Xinyun Chen, Qiuping Huang, Matthew Purver, John R Woodward, Jinxia Xie, and Pengsheng Huang. 2021. Towards robustness of text-to-sql models against synonym substitution. arXiv preprint arXiv:2106.01065.

[8] Ruiqi Zhong, Tao Yu, and Dan Klein. 2020. *Semantic evaluation for text-to-SQL with distilled test suites.* In Proceedings of the 2020 Conference on Empirical Methods in Natural Language Processing (EMNLP), pages 396–411, Online. Association for Computational Linguistics.

[9] Song, D., Schilder, F., Smiley, C., Brew, C., Zielund, T., Bretz, H., Martin, R., Dale, C., Duprey, J., Miller, T., Harrison, J.: TR discover: A natural language interface for querying and analyzing interlinked datasets. Lect. Notes Comput. Sci. (including Subser. Lect. Notes Artif. Intell. Lect. Notes Bioinformatics). 9367, 21–37 (2015). https://doi.org/10.1007/978-3-319-25010-6 2

[10] Saha, D., Floratou, A., Sankaranarayanan, K., Minhas, U.F., Mittal, A.R., Özcan, F.: ATHENA: An Ontology-Driven System for Natural Language Querying over Relational Data Stores. Proc. VLDB Endow. 9, 1209–1220 (2016). https://doi.org/10.14778/2994509.2994536

[12] Sen, J., Lei, C., Quamar, A., Özcan, F., Efthymiou, V., Dalmia, A., Stager, G., Mittal, A., Saha, D., Sankaranarayanan, K.: ATHENA++: natural language querying for complex nested SQL queries. Proc. VLDB Endow. 13, 2747–2759 (2020). https://doi.org/10.14778/3407790.3407858

[13] Baik, C., Arbor, A., Arbor, A., Arbor, A., Jagadish, H. V: Constructing Expressive Relational Queries with Dual-Specification Synthesis. 10th Annu. Conf. Innov. Data Syst. Res. (CIDR '20). (2020).

[14] Dou, L., Gao, Y., Pan, M., Wang, D., Che, W., Zhan, D., & Lou, J.-G. (2023). MultiSpider: Towards benchmarking multilingual text-to-SQL semantic parsing. Proceedings of the AAAI Conference on Artificial Intelligence, 37(11), 12745–12753. https://doi.org/10.1609/aaai.v37i11.26499

[15] Charles T. Hemphill, John J. Godfrey, and George R. Doddington. 1990. The ATIS spoken language systems pilot corpus. In Speech and Natural Language: Proceedings of a Workshop Held at Hidden Valley, Pennsylvania, June 24-27,1990.

[16] John M. Zelle and Raymond J. Mooney. 1996. Learning to parse database queries using inductive logic programming. In AAAI/IAAI, Vol. 2.

[17] Oren Etzioni Ana-Maria Popescu and Henry Kautz. 2003. Towards a theory of natural language interfaces to databases. In Proceedings of the 8th International Conference on Intelligent User Interfaces, pages 149–157.

[18] Zettlemoyer, L.S., Michael, C.: Learning to map sentences to logical form: Structured classification with probabilistic categorial grammars. Proc. 21st Conf. Uncertain. Artif. Intell. UAI 2005. 658–666 (2005).

[19] Zhong, V., Xiong, C., Socher, R.: Seq2Sql: Generating Structured Queries From Natural Language Using Reinforcement Learning. https://arxiv.org/abs/1709. 00103v7. 1–12 (2017).

[20] Qingkai Min, Yuefeng Shi, and Yue Zhang. 2019. A pilot study for Chinese SQL semantic parsing. In Proceedings of the 2019 Conference on Empirical Methods in Natural Language Processing and the 9th International Joint Conference on Natural Language Processing, EMNLP-IJCNLP 2019, Hong Kong, China, November 3-7, 2019, pages 3650–3656. Association for Computational Linguistics.

[21] Anh Tuan Nguyen, Mai Hoang Dao, and Dat Quoc Nguyen. 2020. A pilot study of text-to-sql semantic parsing for Vietnamese. CoRR, abs/2010.01891.

[22] Xu, X., Liu, C., Song, D.: SQLNet: Generating structured queries from natural language without reinforcement learning. https://arxiv.org/abs/1711.04436. 1–13 (2017).

[23] Wang, B., Shin, R., Liu, X., Polozov, O., Richardson, M.: RAT-SQL: Relation-aware schema encoding and linking for text-to-SQL parsers. (2019). https://doi.org/10.18653/v1/2020.acl-main.677.

[24] Lyu, Q., Chakrabarti, K., Hathi, S., Kundu, S., Zhang, J., Chen, Z.: Hybrid ranking network for text-to-SQL. https://arxiv.org/abs/2008.04759 1–12 (2020).

[25] Gao, D., Wang, H., Li, Y., Sun, X., Qian, Y., Ding, B., & Zhou, J. (2023). Text-to-SQL Empowered by Large Language Models: A Benchmark Evaluation. CoRR, abs/2308.15363.

[26] Pourreza, M., & Rafiei, D. (2023). DIN-SQL: Decomposed In-Context Learning of Text-to-SQL with Self-Correction. arXiv preprint arXiv:2304.11015.